\documentclass[10pt,twocolumn,letterpaper]{article}

\usepackage{wacv}              % To produce the CAMERA-READY version
\usepackage{graphicx}
\usepackage{amsmath}
\usepackage{amssymb}
\usepackage{booktabs}
\usepackage{tabularx}
\usepackage[accsupp]{axessibility}  % Improves PDF readability for those with disabilities.

\usepackage[pagebackref,breaklinks,colorlinks]{hyperref}

\usepackage[capitalize]{cleveref}
\crefname{section}{Sec.}{Secs.}
\Crefname{section}{Section}{Sections}
\Crefname{table}{Table}{Tables}
\crefname{table}{Tab.}{Tabs.}

\def\wacvPaperID{16} % *** Enter the WACV Paper ID here
\def\confName{LLVM-AD}
\def\confYear{2024}

\begin{document}

\title{LIP-Loc: LiDAR Image Pretraining for Cross-Modal Localization}

% \author{Sai Shubodh Puligilla\thanks{Corresponding author: \url{p.saishubodh@gmail.com} \\ Project page: \url{https://shubodhs.ai/liploc}\\ Full affiliation: Robotics Research Center, KCIS, IIIT Hyderabad }\\
% % \author{Sai Shubodh Puligilla\\
% IIIT Hyderabad\\
% % {\tt\small firstauthor@i1.org}
% % % To save space, use either the email address or home page, not both
% \and
% Mohd Omama\\
% IIIT Hyderabad\\
% % First line of institution2 address\\
% % {\tt\small secondauthor@i2.org}
% \and
% Husain Zaidi\\
% Microsoft\\
% \and
% Udit Singh Parihar\\
% IIIT Hyderabad\\
% \and
% Madhava Krishna\\
% IIIT Hyderabad\\
% }
\author{
  Sai Shubodh Puligilla \thanks{Corresponding author: \url{p.saishubodh@gmail.com}, Project page: \url{https://shubodhs.ai/liploc}} \thanks{Robotics Research Center, KCIS, IIIT Hyderabad}  \and 
  Mohammad Omama\thanks{University of Texas at Austin} \and 
  Husain Zaidi\thanks{Microsoft} \and
  Udit Singh Parihar\footnotemark[2] \and
  Madhava Krishna\footnotemark[2]
}

\maketitle

%%%%%%%%% ABSTRACT
\begin{abstract}
    Global visual localization in LiDAR-maps, crucial for autonomous driving applications, remains largely unexplored due to the challenging issue of bridging the cross-modal heterogeneity gap. Popular multi-modal learning approach Contrastive Language-Image Pre-Training (CLIP)~\cite{clip} has popularized contrastive symmetric loss using batch construction technique by applying it to multi-modal domains of text and image. We apply this approach to the domains of 2D image and 3D LiDAR points on the task of cross-modal localization. Our method is explained as follows: A batch of N (image, LiDAR) pairs is constructed so as to predict what is the right match between N X N possible pairings across the batch by jointly training an image encoder and LiDAR encoder to learn a multi-modal embedding space. In this way, the cosine similarity between N positive pairings is maximized, whereas that between the remaining negative pairings is minimized. Finally, over the obtained similarity scores, a symmetric cross-entropy loss is optimized. To the best of our knowledge, this is the first work to apply batched loss approach to a cross-modal setting of image \& LiDAR data and also to show Zero-shot transfer in a visual localization setting. We conduct extensive analyses on standard autonomous driving datasets such as KITTI and KITTI-360 datasets. Our method outperforms state-of-the-art recall@1 accuracy on the KITTI-360 dataset by 22.4\%, using only perspective images, in contrast to the state-of-the-art approach, which utilizes the more informative fisheye images. Additionally, this superior performance is achieved without resorting to complex architectures. Moreover, we demonstrate the zero-shot capabilities of our model and we beat SOTA by 8\% without even training on it. Furthermore, we establish the first benchmark for cross-modal localization on the KITTI dataset.
\end{abstract}

%%%%%%%%% BODY TEXT
\section{Introduction}
\label{sec:intro}

Visual localization serves as a crucial aspect in the domain of mobile robotics, playing a pivotal role in applications such as autonomous vehicles and Simultaneous Localization and Mapping (SLAM). Our focus in this work is to address the complex issue of determining the pose of an image within an expansive 3D map. This task of localization gets quite challenging due to large, occluded, dynamic scenes and repetitive places.

Outdoor and indoor visual localization approaches may differ in their pipelines because of the different nature of their challenges; our work addresses outdoor visual localization. However, they can both broadly be categorized into 2 steps: global localization, which gives a rough estimate of the pose, and then, local localization, which gives a more accurate version that further involves the use of PnP~\cite{pnp1, pnp2, pnp3}  in a RANSAC ~\cite{ransac} setting. In our paper, we deal with the problem of global localization. Global localization can be achieved using Global Navigation Satellite Systems (GNSSs), however, it does not always give reliable estimates. This is particularly prevalent in urban environments where high-rise buildings can interfere with the signal quality, leading to inaccuracies. Other contributing factors can include multipath effects, where signals bounce off multiple surfaces before reaching the receiver, and atmospheric conditions, which can alter the signal speed. Therefore, Light Detection And Ranging (LiDAR)-based ~\cite{18-lpdnet,004-pointnvlad} and vision-based approaches ~\cite{001-netvlad,002-end-img-retr,003-deep-feat-vpr} are seen as established sensor modalities in the vision community to estimate the pose of a robot accurately. While LiDAR modality is robust to variation in illumination and can detect objects at long distances with high accuracy, they are generally expensive and are especially prone to failure modules such as degenerate places like tunnels and suffers from issues such as surface reflections and interference. Vision modality-based methods use 2D images and extract features from them using methods such as NetVLAD ~\cite{001-netvlad} to eventually match them with a query image for localization. While vision-based approaches have seen large success, there are still important limitations, such as dynamic environments, illumination, or weather changes. 

Given the complementary nature of these sensor modalities, their advantages can be combined in a multi-modal fashion through the fusion of 2D and 3D data ~\cite{005-oneshot, 006-augment-vpr, 03-adafusion}, which enhances localization accuracy significantly. However, this fusion is not straightforward because of the heterogeneity gap between these two modalities, and therefore, it remains an unexplored and largely unsolved problem. Further, the multi-modal approach does not solve the problem of localizing a sensor of one modality in a map of another, something that we refer to as 'cross-modal localization'. Our task revolves around the novel application of a contrastive loss based on batch construction approach to the distinct domains of 2D images and 3D LiDAR point clouds, specifically in the context of cross-modal localization. This involves creating a shared embedding space for both 2D and 3D data, enabling the localization process to occur even when only one modality is available at a given time.

The practical motivation of our work lies in its utility in autonomous navigation scenarios. Imagine a setting where a detailed and expensive LiDAR map has been constructed using a resource-intensive setup initially. In subsequent navigation instances, however, our approach enables localization solely via 2D images, thereby eliminating the need for resource-heavy operations. This feature is particularly advantageous as it mitigates resource constraints and provides an economical and efficient solution for repeat localizations. Furthermore, our method showcases its versatility by being applicable even when the initial map is constructed in a different modality, such as 3D. Thus, our work facilitates cost-effective and efficient navigation by capitalizing on the power of cross-modal localization.

The main contributions of our paper are as follows:
%-------------------------------------------------------------------------

\begin{itemize}
    \item \textbf{Batched Loss Approach:} This work is the first of its kind to apply the batched contrastive approach in a cross-modal setting involving image and LiDAR data, establishing a novel direction in metric learning for autonomous driving applications.
    \item \textbf{Superior Performance with Simpler Methods:} We demonstrate that our method outperforms state-of-the-art~(\textit{AECMLoc})\cite{02-aecmloc} recall@1 accuracy on the KITTI-360~\cite{kitti-360} dataset by 22.4\% using only perspective images and standard Vision Transformer~\cite{vit_dosovitskiy2021} architecture for the encoders, contrasting with the state-of-the-art approaches that rely on more informative fisheye images and complex architectures.
    \item \textbf{Zero-shot Analyses and Benchmark Establishment:} We conduct exhaustive analyses on standard autonomous driving datasets such as KITTI~\cite{kitti} and KITTI-360~\cite{kitti-360} and establish the first benchmark for cross-modal localization on the KITTI dataset. 
\end{itemize}

The remainder of this paper is organized in the following manner: Section II provides an overview of previous research in the domain of visual localization. Our proposed methodology, encompassing batch construction, contrastive loss, and architecture design, is detailed in Section III. Experiments conducted on public datasets and their results are exhibited in Section IV. In Section V, we demonstrate the zero-shot capability of our model. The paper concludes with Section VI.

%-------------------------------------------------------------------------

\section{Related Work}
\label{sec:formatting}

Localization of a robot involves understanding where it is in the world using a pre-existing map. Generally, this has been done using the same type of sensor that was used to create the map, such as images with images or 3D scans with 3D scans, i.e., between the same corresponding modalities. Our work expands upon this by showing that you can localize using different types of sensors than those used to create the map, like using simple cameras to localize in a map built from expensive 3D scans, which is more flexible and cost-effective. We review the literature on both of these approaches in this section.
%-------------------------------------------------------------------------
\subsection{Same modal localization}

The standard pipeline for localization approaches begins with the acquisition of reference data, which is typically a large 3D map. The first step in the pipeline is to retrieve prior information for which traditional methods include the bag of words approach~\cite{bow}, but more recent work has leveraged deep learning techniques. For instance, Arandjelovic \etal~\cite{001-netvlad} extended the Vector of Locally Aggregated Descriptors (VLAD)~\cite{vlad} approach, introducing a differentiable generalized VLAD layer that can be integrated into any CNN architecture, i.e., NetVLAD~\cite{001-netvlad}, a CNN-based image retrieval algorithm, retrieves the most similar images, or reference images, from an image database. This stage is called global localization, where we can directly take the pose based on the most similar reference images for a given query image. But this pose can be refined further as follows. Once the reference images are retrieved, local feature extraction and matching are performed. Feature extraction has traditionally relied on methods such as Scale-Invariant Feature Transform (SIFT)~\cite{sift}, Speeded Up Robust Features (SURF)~\cite{surf}, and Oriented FAST and Rotated BRIEF (ORB)~\cite{orb}, which are designed to extract local features from the image. However, with the advent of deep learning, recent years have seen a shift towards the use of convolutional neural networks (CNNs) such as ResNet~\cite{resnet}, ConvNets~\cite{alexnetcnn} and more recently, Vision Transformers ViT~\cite{vit_dosovitskiy2021} as a means to extract features. These neural networks have shown remarkable performance in feature extraction, replacing handcrafted features with learned representations. Following local feature matching, the Pose from N Points (PnP)~\cite{pnp1, pnp2} algorithm along with RANSAC is leveraged to estimate the 6-DoF pose. This whole process to obtain finer estimate of pose is termed local localization. In our current work, we apply our batched loss approach to the global localization problem. No single localization method currently exists that can universally adapt to a wide range of environments, such as urban landscapes, rural settings, nighttime conditions, warehouses, varying weather, foggy conditions, and busy marketplaces~\cite{oxford_robotcar, rio10wald2020, cmu_seasons_sat, warehouse_man, inloctaira2018, Cui_2024_WACV}. Most state-of-the-art methods specialize in one or a few of these scenarios and lack the generalizability to perform well across all. Our paper aims to take a step in this direction towards creating a more general localization method.

So far, we have discussed about image-to-image based retrieval methods. Now let us discuss the same modal retrieval for point clouds. Point clouds provide a robust way to represent scenes under varying lighting conditions and seasonal changes and have the ability to maintain the structural integrity of scenes. The initial developments in this field were PointNet~\cite{qi2017pointnet}, and EdgeConv~\cite{edgeconv}, which process unordered points to extract permutation-invariant features. Building upon this, PointNetVLAD~\cite{004-pointnvlad} introduced an end-to-end trainable model, which merges the strengths of PointNet and NetVLAD for point cloud-based place recognition. When a query point cloud is provided, the task involves retrieving the most similar sub-maps from this database, i.e., this is 3D-3D based retrieval. However, the cost and weight of LiDAR technology can be prohibitive for large-scale applications. Here is where our work has significant motivation. Consider a scenario where an exhaustive and expensive map has been previously constructed using a combination of sensor setups. With our approach, you can localize within this map multiple times using only simple and cost-effective sensors, such as RGB cameras. This eliminates resource constraints and allows for flexibility even if the map has not been pre-built in that specific sensor modality. 

\begin{figure}[t]
  \centering
    \includegraphics[width=\linewidth]{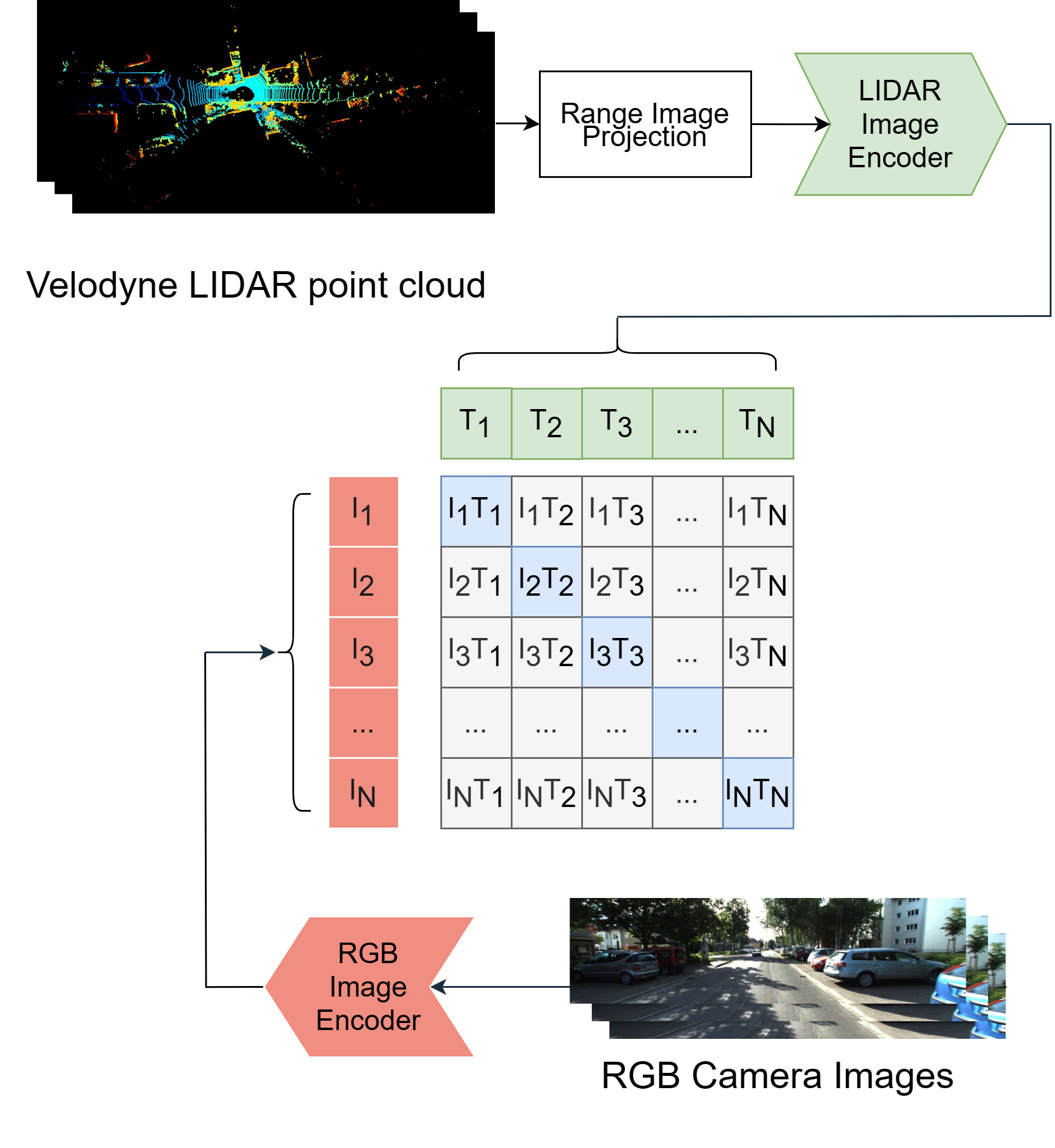}
   \caption{Batched Contrastive Learning Architecture}
   \label{fig:onecol}
\end{figure}

%-------------------------------------------------------------------------
\subsection{Cross modal localization}

Cross modal localization has gained traction in recent years \cite{07-cmrnet, 08-lccnet, 09-pwcnet, omama2023alt}. Early works which localize between the domains of LiDAR and 2D images include ~\cite{07-cmrnet, 08-lccnet, 09-pwcnet}. These approaches use CNN based approaches to do pose regression with standard translation or rotation-based losses. These methods typically need some initial estimate to work and only works as a local localization task. 2D3D-MatchNet~\cite{05-matchnet} proposes a deep network to jointly learn the 2D image and 3D point cloud keypoint descriptors. 

More recently, Cattaneo \etal ~\cite{01-global} proposed a teacher-student training approach on Oxford Robotcar using triplet loss and jointly trains a 2D network for the images and 3D network for the point clouds. Similarly, AdaFusion~\cite{03-adafusion} presents a visual-LiDAR descriptor fusion in a weighted way using a pairwise margin-based loss. A similar approach i3dLoc~\cite{04-i3dloc} uses Generative Adversarial Networks (GANs) on the domains of 360-images and 2D range projections to deal with inconsistent environmental conditions on a custom dataset. Recently, LiDARCLIP~\cite{hess2023lidarclip} uses principles from CLIP to produce joint text, LiDAR and camera embeddings, although not for the task of global localization. The work closest to ours is AECMLoc~\cite{02-aecmloc} which is the first work to address the task of cross-modal localization on KITTI-360 dataset. It uses a spherical convolution based 2D network and another PointNet based 3D network with attention enhancement with a triplet loss and achieves reasonable cross-modal localization accuracy. 
In contrast to traditional methods which employ a triplet loss with select negative samples through hard negative mining, we introduce the use of a contrastive loss incorporating numerous negative samples within a batch construction method. 

%-------------------------------------------------------------------------

\section{Methodology}
\label{sec:formatting}

In this section, we discuss in detail our methodology where we first discuss how the data is constructed in a batched manner and how contrastive loss is imposed upon it. Then we discuss the details about encoder architecture and training.
%-------------------------------------------------------------------------
\subsection{Batch Construction and Contrastive Loss}

Here, we discuss the batch construction procedure and the contrastive symmetric loss in detail. This approach was first introduced as multi-class N-pair loss~\cite{NIPS2016_batched_loss} and then popularized by InfoNCE~\cite{loss_clip_nce} and CLIP~\cite{clip} under various names. In the context of our paper, we call it batched contrastive loss.

Firstly, our batch construction method is explained as follows: A batch of $N$ (image, LiDAR) pairs is constructed so as to predict what is the right match between $N X N$ possible pairings across the batch by jointly training an image encoder and LiDAR encoder to learn a multi-modal embedding space. In this way, we would have $N$ positive pairings and $(N^2 - N)$ negative pairings as shown in Fig \ref{fig:onecol}.

To put it formally, let us say  $x_i$ represents a 2D image, $x_i^{+}$ represents a LiDAR sample, and $f(x)$ represents an embedding vector for $x$, simply written as $f$. In a batch size of $N$, we have $N$ such pairs $\left\{\left(x_1, x_1^{+}\right), \cdots,\left(x_N, x_N^{+}\right)\right\}$. First, let us consider for one 2D image (the same explanation works vice-versa for LiDAR given the symmetric nature of loss). Consider an $(N+1)$-tuplet of one 2D image and $N$ LiDAR samples i.e. $S_i=\left\{x_i, x_1^{+}, x_2^{+}, \cdots, x_N^{+}\right\}:$  The anchor here is $x_i$ while $x_i^{+}$ is a positive example to the anchor and $x_j^{+}, j \neq i$ are the negative examples. In other words, every 2D image would have 1 positive and $N-1$ negative LiDAR examples (and vice-versa). Therefore, our loss can be finally expressed as:

\begin{align}
\mathcal{L}_{\text {batched}}\left(\left\{\left(x_i, x_i^{+}\right)\right\}_{i=1}^N ; f\right) &= \frac{1}{N} \sum_{i=1}^N \log \Bigg(1 \nonumber \\
&+ \sum_{j \neq i} \exp \left(f_i^{\top} f_j^{+}-f_i^{\top} f_i^{+}\right)\Bigg)
\label{eq:batched-loss}
\end{align}

Each $i$ in the outer summation would correspond to every row in the \cref{fig:onecol}. The above loss can equivalently be expressed as standard softmax loss as follows (for full theoretical details, refer to ~\cite{NIPS2016_batched_loss}):

% \begin{align}
% \mathcal{L}_{\text {batched}}\left(\left\{\left(x_i, x_i^{+}\right)\right\}_{i=1}^N ; f\right) \nonumber \\
% = \frac{1}{N} \sum_{i=1}^N \Bigg\{ &\log \left(\frac{\exp \left(f_i^{\top} f_i^{+}\right)}{\exp \left(f_i^{\top} f_i^{+}\right) \right. \nonumber \\ 
% &\left. + \sum_{j \neq i} \exp \left(f_i^{\top} f_j^{+}\right)}\right) \Bigg\}
% \label{eq:batched-loss-3}
% \end{align}

\begin{equation}
\begin{split}
\mathcal{L}_{\text {batched}}\left(\left\{\left(x_i, x_i^{+}\right)\right\}_{i=1}^N ; f\right) = \\
-\frac{1}{N} \sum_{i=1}^N \log \left(\frac{\exp \left(f_i^{\top} f_i^{+}\right)}{\exp \left(f_i^{\top} f_i^{+}\right)+\sum_{j \neq i} \exp \left(f_i^{\top} f_j^{+}\right)}\right)
\end{split}
\label{eq:batched-loss-3}
\end{equation}

This loss is used to train our dual encoders as explained in the next section.

\begin{figure}[t]
  \centering
    \includegraphics[width=\linewidth]{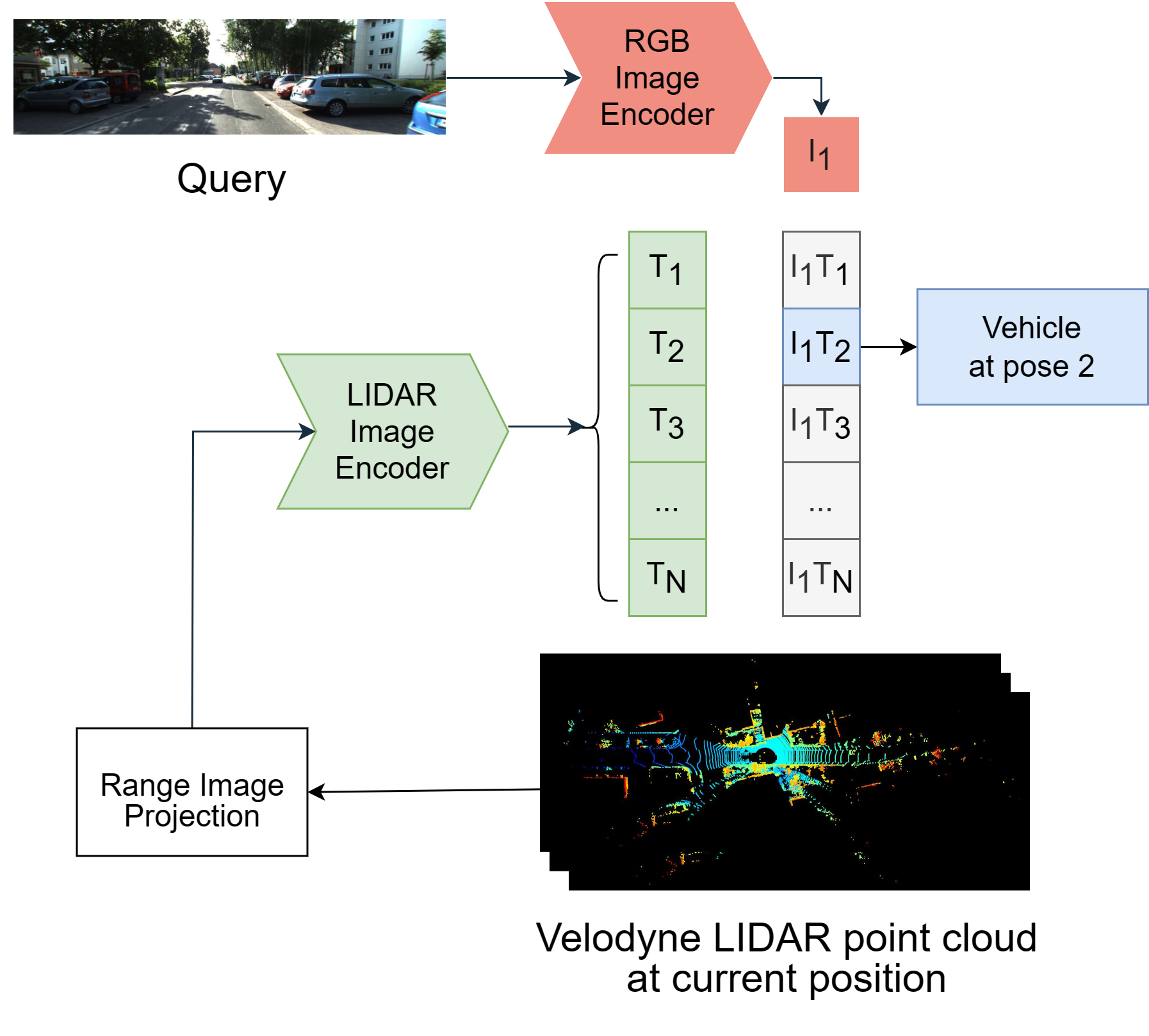}
   \caption{Inference pipeline showing that query camera image is calculating similarity with all the lidar range images in the database}
   \label{fig:twofig}
\end{figure}

\begin{figure}[t]
  \centering
    \includegraphics[width=\linewidth]{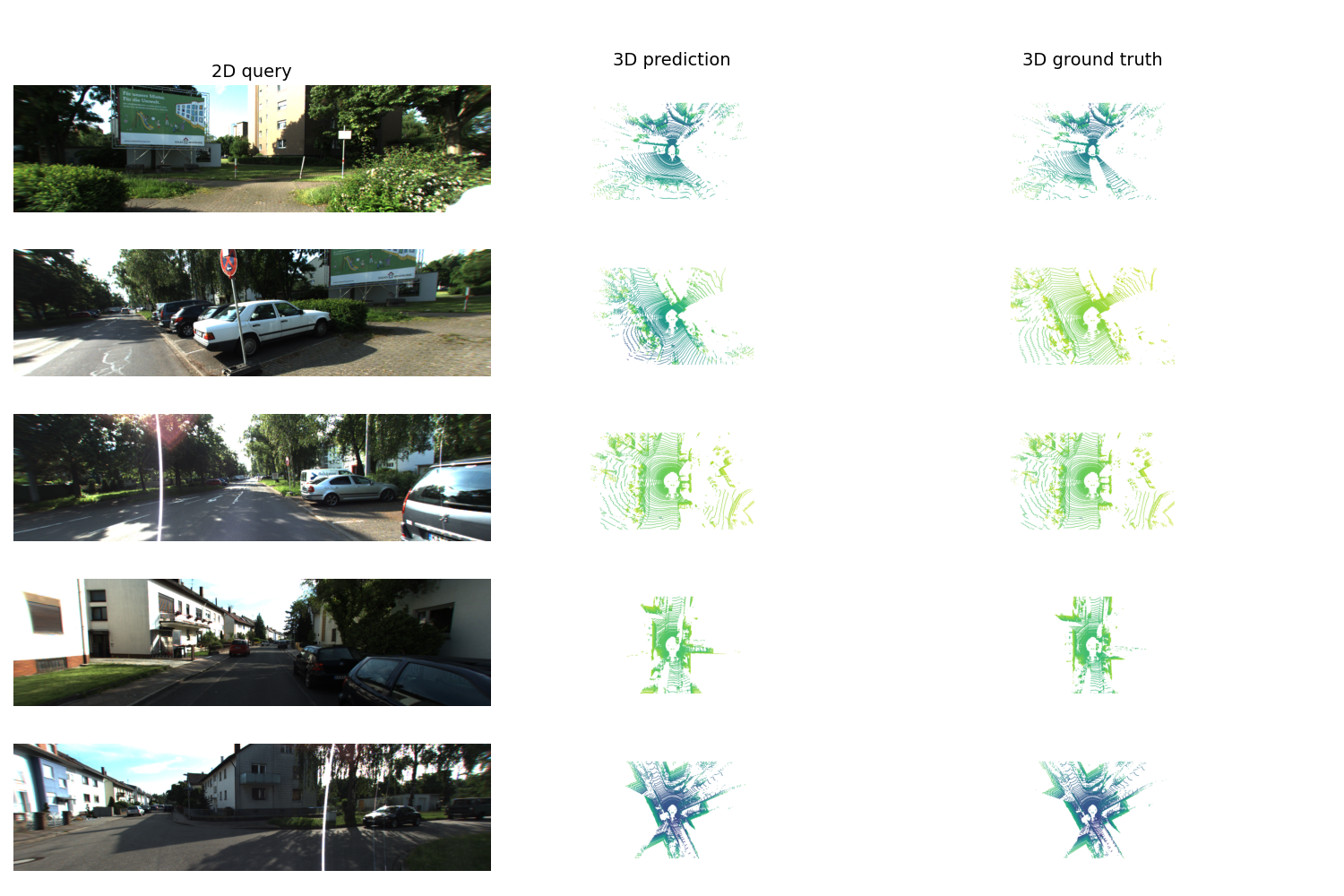}
   \caption{Qualitative visualization: 2D query, 3D prediction, 3D ground truth.}
   \label{fig:threefig}
\end{figure}
%-------------------------------------------------------------------------
\subsection{Contrastive encoder training}

Our methodology uses contrastive learning to jointly train two encoders: one for 2D images from the camera and the other for 3D points from the LiDAR. We generate range images from the LiDAR points. In our experimentations, we found that using range images when passed through standard ResNet or ViT architectures gave us better performance in comparison with specialized or sophisticated architectures, which use 3D point clouds as input. The training process ensures corresponding camera and LiDAR images are closely aligned by learning a shared multi-modal embedding space and, thus, enables efficient cross-modal localization.

Both encoders follow the same two-step process: feature extraction and projection. For feature extraction, we employ the established ViT~\cite{vit_dosovitskiy2021} model. We have also experimented with other models such as ResNet~\cite{resnet} which too gives superior performance to baseline (ResNet backbone beats baseline by more than 10\%), however amongst these two, ViT is much better. We deliberately opted for a standard architecture, aiming to highlight the effectiveness of our batched loss method without resorting to more advanced architectures. The feature extraction step uses the pre-trained \textit{vit\_small\_patch16\_224} model to transform the input images into intermediate feature vectors, and then, the feature projection step transforms these high-dimensional vectors into a shared and lower-dimensional embedding space. The projection step is aimed at bolstering the task-specific features and preserving the compact nature of embeddings to ensure computational efficiency. We found \textit{vit\_small\_patch16\_224} to be our best model amongst other variants, henceforth, whenever not explicitly mentioned, it can be assumed that we are referring to this model by 'LIP-Loc'.

The joint training of the encoders ensures both the camera and LiDAR encoders update their parameters during each training step simultaneously to maximize the cosine similarity between the embeddings of the corresponding camera and LiDAR pairs and to minimize the similarity between those of non-corresponding pairs. This is achieved by optimizing a symmetric cross-entropy loss based on a batch construction procedure, as explained previously. 

%-------------------------------------------------------------------------

\section{Experiments and Results}
\label{sec:formatting}

We have shown the effectiveness of our contrastive learning via Batch Construction in two different datasets, KITTI and KITTI 360. Our ablation results show the improvement of batched contrastive loss over triplet loss in terms of network recall and GPU memory footprint. With the incorporation of cropping in the image field of view and distance thresholding in LiDAR space, we are able to achieve better generalization, as shown quantitatively in the table. 
%-------------------------------------------------------------------------
\subsection{Dataset and Preprocessing}

For both KITTI and KITTI 360, our training pairs consist of synchronized LiDAR scans and camera images taken in the same snapshot of the world.

\textbf{KITTI dataset}

Evaluation is performed on the KITTI odometry dataset which consists of Velodyne HDL-64E LiDAR scans and a color stereo camera rig. Our experiment id and corresponding training sequences are shown in Table \ref{tab:training_seqs}. The evaluation sequences are 08 and 09.

\begin{table}
  \centering
  \begin{tabular}{@{}lc@{}}
    \toprule
    Experiment id & Training sequences \\
    \midrule
   exp\_large   & KITTI(5): 03, 04, 05, 06, 07 \\
   exp\_larger  & KITTI(8): 00, 01, 02, 03, 04, 05, 06, 07 \\
   exp\_largest & KITTI(18): 00, 01, 02, 03, 04, 05, 06, 07, 11,\\ 
               & 12, 13, 15, 16, 17, 18, 19, 20, 21 \\
    exp\_360 & KITTI\_360(7): 03, 04, 05, 06, \\ 
               & 07, 09, 10 \\
    \bottomrule
  \end{tabular}
  \caption{Training sequences for different experiments.}
  \label{tab:training_seqs}
\end{table}

\textbf{KITTI 360 dataset}

It comes with Velodyne HDL-64E and raw images from the perspective camera. This has about 80k frames of lidar and image pairs over a distance of 73.7 Km, along with precise vehicle pose information require for evaluation. We have used sequences 3,4,5,6,7,9,10 for training and sequence 0 for evaluation. Of course, in the Zero-shot setting, we did not use any of this. To not confuse the standard train-test evaluation with Zero-shot, we separately dedicate Section 5 for Zero-shot results. We have applied distance threshold in lidar scans and field of view threshold on lidar range image.

%-------------------------------------------------------------------------
% \subsection{Evaluation metrics}
% \begin{figure}[t]
%   \centering
%   % \fbox{\rule{0pt}{2in} \rule{0.9\linewidth}{0pt}}
%     \includegraphics[width=\linewidth]{InferencePipeline.png}
%    \caption{Inferring with query}
%    \label{fig:twofig}
% \end{figure}

% \begin{figure}[t]
%   \centering
%   % \fbox{\rule{0pt}{2in} \rule{0.9\linewidth}{0pt}}
%     \includegraphics[width=\linewidth]{ExampleImages.png}
%    \caption{RGB and corresponding LIDAR Images}
%    \label{fig:threefig}
% \end{figure}

\subsection{Objective and Evaluation metrics}

We further clearly elaborate the objective here. During training time, we take a batch of N (image, LiDAR) pairs, where N is typically 32 as illustrated in Fig \ref{fig:onecol} which we train for 50 number of epochs.  It is hereby reiterated that hyperparameter tuning is not necessary and our method works out of the box as mentioned in next section in even a zero shot setting. The inference procedure is explained in Fig \ref{fig:twofig}, where given an RGB query, we pass it through its corresponding encoder to extract the embedding and similarly for reference LiDAR range images to extract corresponding LiDAR embeddings. Then, the reference with highest similarity score with query is picked as final prediction whose pose is finally extracted for the global localization problem. The same process is done vice-versa for 3D to 2D localization task.

We used the Recall@1 evaluation metric with a distance threshold of 20 meters. For a given query camera image embedding we find the closest LiDAR range image embedding in joint space, if the distance between the retrieved camera and LiDAR is less than 20 meters, then we consider it as a True Positive. Thus,

\[
\text{Recall@1} = \frac{\text{Number of True LiDAR and camera matches}}{\text{Total number of query camera images}}
\]

%-------------------------------------------------------------------------
\subsection{Comparative Analysis: Triplet Versus Batched Contrastive}

Our experimental findings align with the premise that an increase in negative samples in metric learning contributes to better generalization, a claim also made by foundational works such as CLIP \cite{clip}. We observe the same in the results we obtained, as illustrated in Table \ref{tab:triplet_vs_batch_contrastive}.

The traditional triplet approach tends to require more GPU memory as the number of negative samples increases \cite{NIPS2016_batched_loss}. This can create limitations when working with larger datasets or when more robust learning is required. 

In contrast, the batched contrastive approach presents a more efficient and scalable solution. It has demonstrated improved performance with the addition of more data \cite{NIPS2016_batched_loss}, making it particularly suitable for metric learning. The scalability of this approach allows for the management of larger datasets, making it a key consideration for improving the efficiency of intermediate modules in the learning process.
 Table \ref{tab:triplet_vs_batch_contrastive} shows comparison of the batched contrastive loss against standard triplet loss for lidar-camera alignment on the KITTI dataset. Please note that we use ResNet50 for this analysis and not ViT. We see  that with batched construction, we can get more negative samples with lesser GPU footprint leading to better evaluation results. 

\begin{table}[h]
\centering
\begin{tabular}{|l|l|l|}
\hline
Exp\_ID/Metric & triplet loss & batched loss \\
\hline
Seq 8 r@1 & 0.215 & 0.295 \\
\hline
Seq 9 r@1 & 0.232 & 0.309 \\
\hline
GPU Utilization & $\sim$8214MB$\times$2 & $\sim$8214MB \\
\hline
No of -ve Samples & 1 & 31 \\
\hline
Batch Size & 32 & 32 \\
\hline
\end{tabular}
\caption{Recall@1 comparison between Triplet and Batched Contrastive Training for \textit{exp\_larger} setting i.e. no of sequences is 8.}
\label{tab:triplet_vs_batch_contrastive}
\end{table}

%-------------------------------------------------------------------------
\subsection{Data Preprocessing for Better Generalization}

A lower quantity of data can potentially reduce the model's ability to generalize effectively. To mitigate this, we advocate the incorporation of intelligent pre-processing techniques designed to boost generalization. In our study, we experiment with distance threshold cropping for LiDAR data and field of view (FoV) cropping for LiDAR range images. The horizontal field of view of a LiDAR range image is greater than the camera field of view, so we crop the LiDAR range image such that both sensors have common overlapping information present.

The rationale for utilizing distance cropping stems from the observation that objects at a greater distance while being accurately captured by LiDAR, may not be equally discernible through camera imaging. In contrast, if we constrain the LiDAR data too much to a more confined area, we risk losing sight of more distant meaningful, and relevant information for our model.

Our empirical analysis in Table \ref{tab:lidar_crop} strongly indicates that a distance threshold of 50 meters for LiDAR cropping provides the most beneficial outcome for our model's performance. Adjusting this threshold to either higher or lower distances tends to degrade the overall results, underscoring the critical role of this specific parameter in optimizing the data preprocessing stage for better generalization.

\begin{table}[h]
\centering
\begin{tabular}{|l|l|l|l|}
\hline
Sequence/ExpID & Seq 8 & Seq 9\\
\hline
exp\_larger $\|$ No Threshold & 0.295 & 0.309\\
\hline
exp\_larger $\|$ (50m Threshold) & 0.325 & 0.370\\
\hline
exp\_larger $\|$ (30m Threshold) & 0.317 & 0.308\\
\hline
exp\_largest $\|$ No Threshold & 0.484 & 0.457\\
\hline
exp\_largest $\|$ (50m Threshold) & 0.540 & 0.495\\
\hline
\end{tabular}
\caption{Ablation of thresholding on lidar scans with increase in training sequences}
\label{tab:lidar_crop}
\end{table}

%-------------------------------------------------------------------------
\subsection{Scaling Data for Better Generalization}

The robustness of metric learning is directly proportional to the volume of data available for processing\cite{clip}. This concept is clearly exemplified in our empirical results, as presented in Table \ref{tab:dataset_scaling}. As we progressively increase the number of sequences, there is a noticeable upswing in accuracy across both sequences.

To conclude, our results emphasize the importance of leveraging scalable techniques and large, varied datasets in metric learning, as this approach can notably enhance model accuracy and its generalization capabilities.

\begin{table}[h]
\centering
\begin{tabular}{|l|l|l|l|}
\hline
 Sequence/ExpID & Seq 8 & Seq 9 & Sequences \\
\hline
exp\_large & 0.278 & 0.260 & 5 \\
\hline
exp\_larger & 0.547 & 0.525 & 8 \\
\hline
exp\_largest & 0.805 & 0.780 & 18 \\
% \hline
% exp\_combined & 0.705 & 0.788 & 18 KITTI + \\
%               &       &       & 7 KITTI-360 \\
\hline
\end{tabular}
\caption{Increasing in number of sequences and mixture of datasets leads to better generalization}
\label{tab:dataset_scaling}
\end{table}

%-------------------------------------------------------------------------
\subsection{Setting New Benchmark on KITTI-360 Cross-Modal Place Recognition Task}

% In this section, we demonstrate that our approach surpasses the previous state-of-the-art \textit{AECMLOC}\cite{02-aecmloc} results on the KITTI-360 dataset, as preseneted in Table \ref{tab:aecmloc_2d_3d} and Table \ref{tab:aecmloc_3d_2d}. Interestingly, the prior SOTA strategy was reliant on fisheye images, which offer a wider field of view but require specialized preprocessing due to their inherent distortion. In contrast, our approach bypasses the need for such images and instead utilizes perspective images, which are more straightforward to handle and still provide sufficient information for our task. Moreover, the prior SOTA methodology necessitated a specialized spherical convolutional network architecture for processing the fisheye images. In contrast, our approach employs a simple yet powerful ResNet50 architecture coupled with our batched contrastive learning strategy, contributing to its scalability and robustness. Figure \ref{fig:aecmloc_liploc} shows that LIPLoc outperforms AECMLoc in both Zero shot and same dataset training. We have compared across different values \textit{k} for recall. Also for full ablation we have compared both 2D query and 3D database and 3D query and 2D database for recall.

% Furthermore, we observe that the model trained on a combined dataset of KITTI and KITTI-360 significantly outperforms a model trained exclusively on the KITTI-360 dataset, underscoring yet again the value of diverse and extensive datasets in improving model performance.

\begin{figure}[t]
  \centering
    \includegraphics[width=\linewidth]{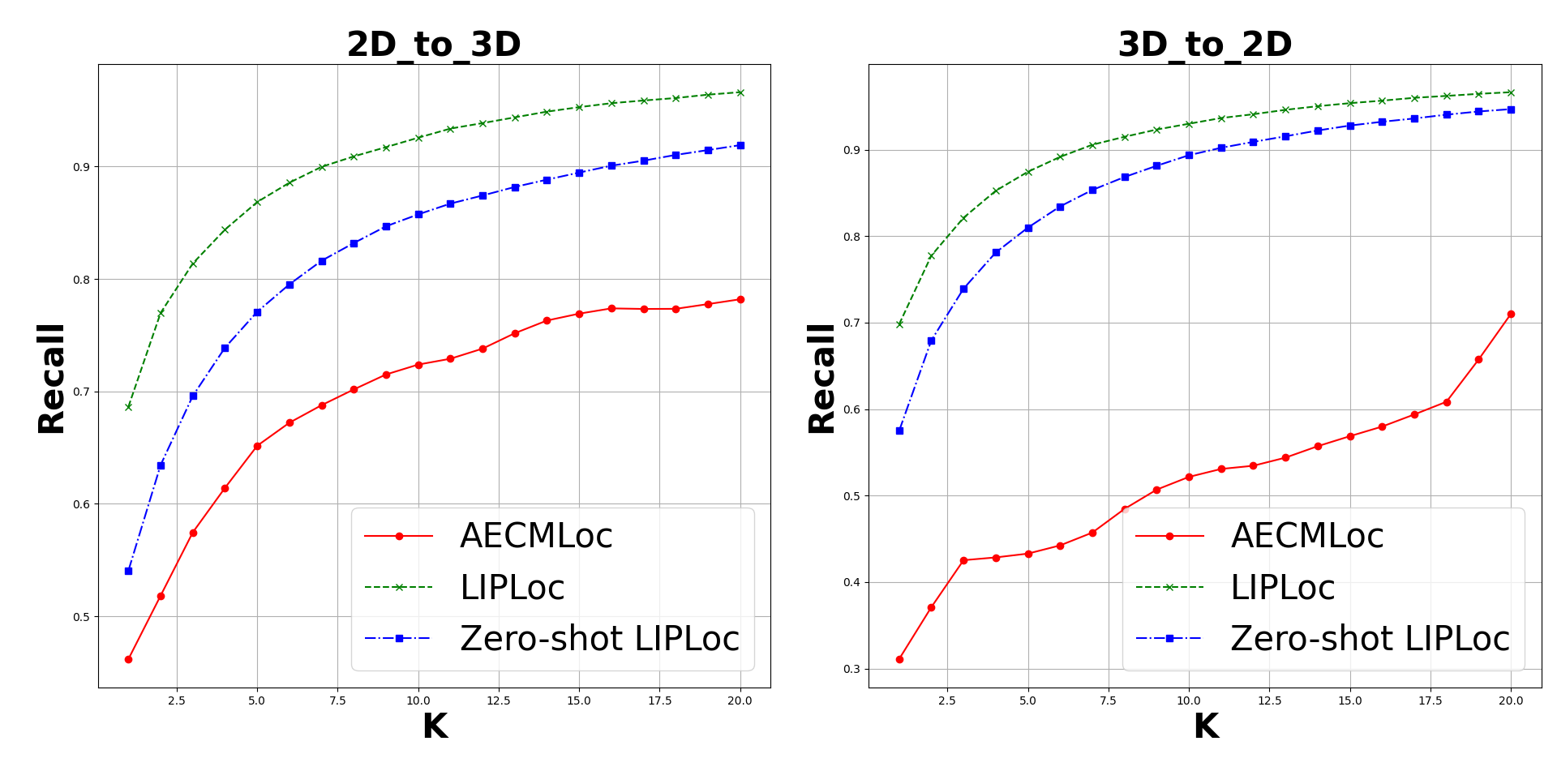}
   \caption{Recall@k plot for various values of k for baseline comparison with our LIP-Loc and Zero-shot LIP-Loc.}
   \label{fig:aecmloc_liploc}
\end{figure}

In this section, we demonstrate our approach's superiority to the prior state-of-the-art \textit{AECMLoc}\cite{02-aecmloc} on the KITTI-360 dataset (Tables \ref{tab:aecmloc_2d_3d} and \ref{tab:aecmloc_3d_2d}). Unlike the former, which relied on fisheye images requiring complex preprocessing due to distortion and uses a specialized architecture, we employ perspective images. Figure \ref{fig:aecmloc_liploc} demonstrates LIPLoc's superiority over AECMLoc in both Zero-shot and standard same-dataset training by plotting recall values from \textit{k}'s value of 1 to 20. LIP-Loc overwhelmingly beats the baseline by about 20\% in both 2D-3D as well as 3D-2D settings; and at Recall@20, we almost reach 97\% accuracy, meaning that our method would be a robust retrieval method for further fine grained localization. In their paper, AECMLoc show that these kind of 95+\% recall@20 values are observed only in same modal localization, i.e. 2D to 2D or 3D to 3D. It is interesting to note that our method reaches similar values while being a cross modal method. Furthermote, notice that baseline's 3D to 2D reduces by 15\% compared to 2D to 3D, whereas ours is almost the same, demonstrating the versatility of our method.
Note that we beat the baseline method without even training on KITTI-360 by 8\%. The next section addresses this in detail.

\begin{table}[htbp]
\centering
\begin{tabular}{|c|c|c|c|}
\hline
2D to 3D & recall@1 & recall@5 & recall@20 \\
\hline
AECMLoc & 0.462 & 0.660 & 0.782 \\
\hline
LIP-Loc & 0.686 & 0.868 & 0.966 \\
\hline
Zero-shot LIP-Loc & 0.540 & 0.770 & 0.919 \\
\hline
\end{tabular}
\caption{Baseline comparison of Recall values for 2D query to 3D database localization}
\label{tab:aecmloc_2d_3d}
\end{table}

\begin{table}[htbp]
\centering
\begin{tabular}{|c|c|c|c|}
\hline
3D to 2D & recall@1 & recall@5 & recall@20 \\
\hline
AECMLoc & 0.311 & 0.472 & 0.710 \\
\hline
LIP-Loc & 0.6982 & 0.8745 & 0.9665 \\
\hline
Zero-shot LIP-Loc & 0.574 & 0.809 & 0.946 \\
\hline
\end{tabular}
\caption{Baseline comparison of Recall values for 3D query to 2D database localization}
\label{tab:aecmloc_3d_2d}
\end{table}

% \begin{table}[h]
% \centering
% \begin{tabular}{|l|l|}
% \hline
% Exp\_ID & Seq 00 \\
% \hline
% \textit{AECMLOC} & 0.46 \\
% \hline
% exp\_kitti\_360  & 0.545 \\
% \hline
% exp\_combined & 0.637 \\
% \hline
% \end{tabular}
% \caption{Recall@1 in KITTI-360 and mixture of datasets network performance}
% \label{tab:kitti_360_combined}
% \end{table}

%-------------------------------------------------------------------------

\section{Zero-shot Transfer}
\label{sec:zeroshot}
%-------------------------------------------------------------------------
\subsection{Standard Definition of Zero-shot}

Zero-shot learning in the context of computer vision refers to the ability of a model to generalize to classes not seen during training \cite{unseen-zero-lam}. CLIP redefines this term and extends it to refer to generalisation to unseen datasets.

CLIP attempts to emphasize the task-learning capabilities of models through zero-shot transfer; however, since popular computer vision datasets are inclined towards generic image classification rather than task-specific evaluations, their analysis on these datasets primarily serve as assessment to domain generalization and robustness to distribution shift. We also focus on the latter in our paper i.e. domain generalization.

%-------------------------------------------------------------------------
\subsection{Zero-shot Transfer for Localisation}

In the context of visual localization, we define zero-shot transfer as the model's capability to estimate the pose of an object in unseen datasets. Note that as an early work in this area, we are referring to coarse estimate of pose i.e. global localization problem. To the best of our knowledge, zero-shot transfer has never been applied to visual localization previously.

It is worth mentioning that while our work is inspired by CLIP’s application on computer vision tasks, CLIP itself was inspired by GPT-1 \cite{gpt1-radford2018improving} and GPT-2 \cite{gpt2-Radford2019LanguageMA} which have studied zero-shot transfer over the course of pre-training and “unexpected” task-learning capabilties of language models.

In our paper, we train the model on KITTI dataset and evaluate it on KITTI-360, the baseline for which is AECMLoc which is exclusively trained on KITTI-360. Both these datasets are captured from different camera modalities: KITTI-360 employs 180-degree FOV fisheye cameras, while KITTI uses 90-degree FOV cameras. In fact, when KITTI-360 input data is converted to perspective images, there is heavy perspective distortion because of which training directly on perspective results in under-performance as AECMLoc has reported. More importantly, KITTI 360 has no trajectory overlap with KITTI. All these factors make KITTI-360 an interesting candidate for out-of-distribution dataset. We do acknowledge that KITTI-360 might not be analogous to how ImageNet Rendition \cite{imagenet-ren-hendrycks2021faces} or ImageNet Sketch \cite{imagenet-sketch-wang2019learning} was for ImageNet, but all these factors could make it equivalent to ImageNetV2 \cite{imagenet-v2-recht2019imagenet}.

\begin{figure}[t]
  \centering
    \includegraphics[width=\linewidth]{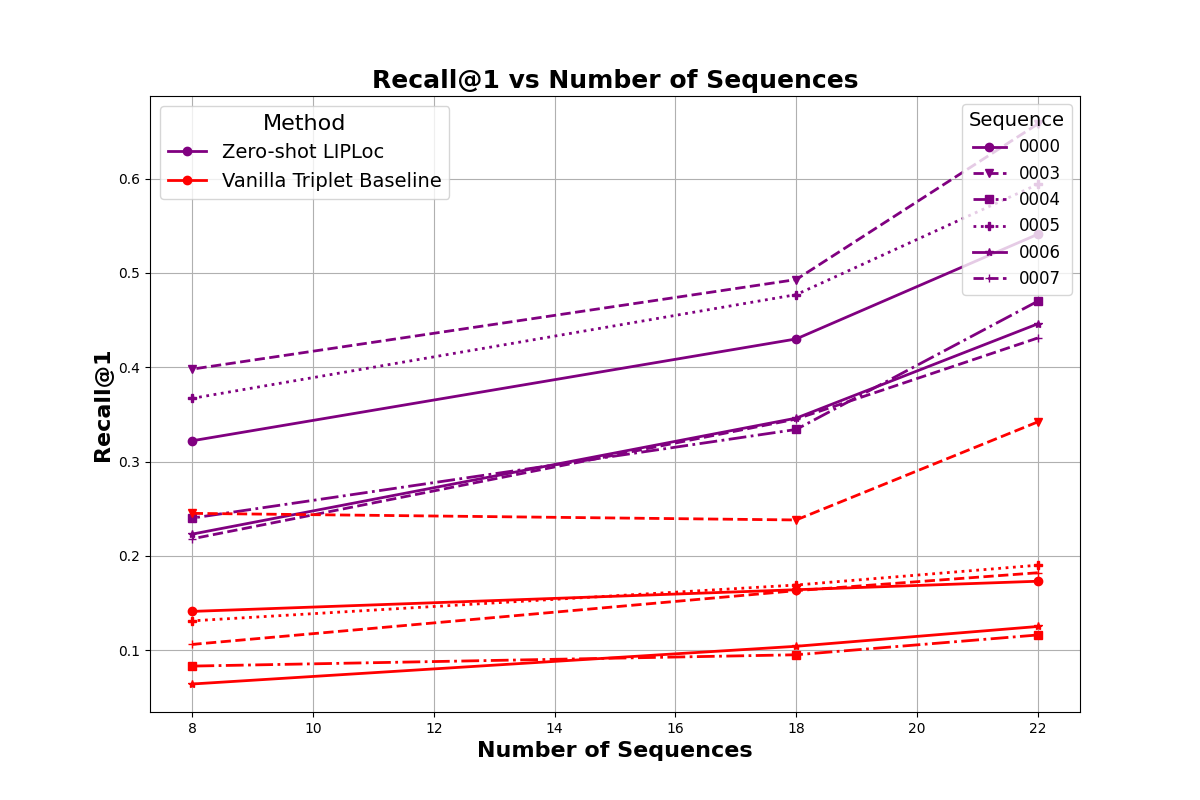}
   \caption{Zero-shot Scaling: Here, x-axis corresponds to the number of sequences of KITTI. y-axis refers to recall@1 on different sequences of KITTI-360. Note that our batched loss scales with increase in dataset size, while triplet loss which most standard methods use shows very marginal increase.}
   \label{fig:zero_shot_scaling}
\end{figure}

%-------------------------------------------------------------------------
\subsection{Results and Analysis: Scaling \& Robustness}

To be truly Zero-shot, we report our accuracy values without doing any customization of hyperparameters for KITTI-360 nor do we form any validation sets on KITTI-360. The trained model truly works out of the box and outperforms AECMLoc which is trained on KITTI-360. We further compare the Zero-shot capabilities of our approach with the 'vanilla-triplet' baseline which uses the standard triplet loss formulation which most standard methods use.

Our best model which we refer to as “Zero-shot LIPLoc” is trained on full KITTI dataset and beats AECMLoc by 8\%. It reaches comparable accuracy to AECMLoc when trained on about 80\% of the KITTI sequences.

It is common in computer vision that models scale with dataset size. However, such phenonemon is non-trivial in localization and hence we have shown in the previous section that increase in number of sequence increases accuracy. In fact, it is specific sometimes to such an extent that there could be a separate model fitted every sequence of dataset \cite{rio10-dong2020robust} and these models don’t even work on other sequences of the same dataset, let alone on another dataset.

Therefore, we go one more step ahead and show in Figure \ref{fig:zero_shot_scaling} that as we scale up training on original KITTI data, the accuracy on KITTI-360 progressively improves although Zero-shot LIP-Loc has never seen KITTI-360; demonstrating domain generalization. The average increase in Zero-shot LIP-Loc is 23\% whereas the that in baseline is 6\%, clearly showing the scalability of constrastive formulation. The more common triplet formulation which is used in most standard methods only increases by few percentage points. We have tried bigger ViT models as encoders, but the accuracy saturates. This is because we are dealing with much smaller datasets, the model may become overparameterized.

In CLIP, as they train on internet scale data, they admit that it is ambiguous what exactly results in accuracy increase: data, model or loss function? But in our case, we exploit the benefit of working with smaller dataset and clearly explained how each factor contributed to our training.

How well does a Zero-shot model work on out of distribution datasets? Typically in deep learning when models are trained and evaluated on same dataset like ImageNet, they exploit spurious correlations because of which robustness gap arises. CLIP does this robustness analysis by testing on 7 natural distribution shift datasets \cite{7-nat-taori2020measuring}. Such a benchmark does not exist for visual localization, therefore our work motivates the building of such a benchmark. 

\begin{figure}[t]
  \centering
    \includegraphics[width=\linewidth]{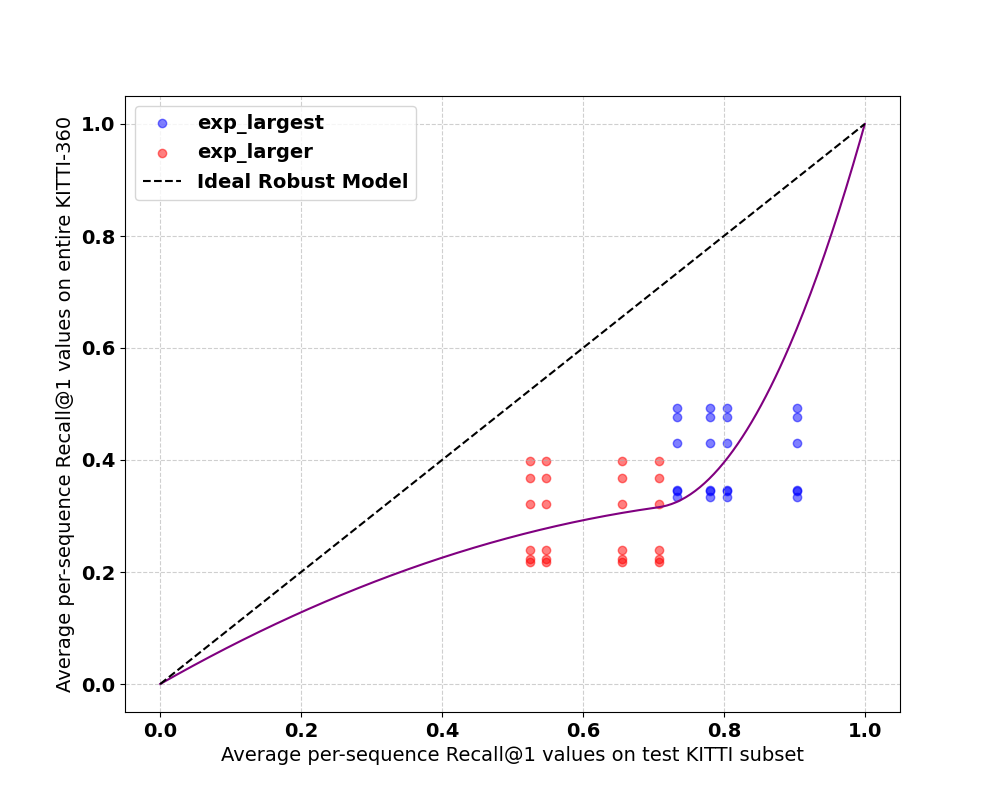}
   \caption{Zero-shot Robustness: When recall on KITI subset is around 50\%, our zero-shot model reaches upto 40\% on KITTI-360; when it is around 80\%, our model reaches upto 50\% recall, strongly demonstrating robustness.}
   \label{fig:zero_shot_robustness}
\end{figure}

Within the scope of this paper, we do the robustness analysis between sequences of KITTI and KITTI-360 in Fig \ref{fig:zero_shot_robustness}. We firstly consider models which are trained on subset of KITTI data as per Table \ref{tab:training_seqs}. Then we evaluate on the rest of KITTI data as test dataset whose recall we plot on x-axis and then evaluate on full KITTI-360 sequences whose recall we plot on y-axis. An ideal robust model would perform equally well on both these test sets, i.e. y=x line. When our curve is closest to the robust model plot, we notice in Fig \ref{fig:zero_shot_robustness} that when there is only about 10-20\% accuracy gap between seen validation dataset and unseen data, proving the robustness of LIP-Loc. This holds true even when the model is trained on a smaller dataset.

Note that this is only the first step towards establishing robustness of Zero-shot localization models. There needs to be a dedicated benchmark along with standardised metrics to truly evaluate these models. Since there is no equivalent of internet scale (text, image) data for localization, how we train our localization models and evaluate them is an open question, as we discuss in the next section.

%-------------------------------------------------------------------------

\section{Future Work}
\label{sec:futurework}
Our approach showcases the potential of batched contrastive learning in bridging the cross-modal heterogeneity gap. By achieving superior performance without relying on complex architectures or fisheye images, our method offers a simpler yet highly effective solution for cross-modal localization. Additionally, we establish the first benchmark for cross-modal localization on the KITTI dataset, providing a foundation for future research.

Our reflections on Zero-shot transfer open a set of thought provoking questions: What is the internet scale equivalent for localization? Can a larger version of dataset like KITTI act as like one? Or would it involve synthetic dataset? What is the equivalent of natural distribution scale datasets for localization? Would recall@K be the right metric for evaluating such systems or do we need better metrics? One of the weaknesses of CLIP is its task learning capabilities, for instance CLIP struggles to find out the closest objects in an image. Could combining depth encoder  with text encoder solve this problem and other task generalization problems?

All of these are predominantly open questions which we hope our work will motivate the readers to address in their own work.

%%%%%%%%% REFERENCES
{\small
\bibliographystyle{ieee_fullname}
\bibliography{egbib}
}

\clearpage

%%% SUPPLEMENTARY
\setcounter{page}{11}
\part*{Supplementary Material} % Creates a part title without a number
\appendix % Resets section numbering and changes it to alphabetical
\section{Qualitative Results \& Semantic Breakdown of KITTI360 }
\label{sec:intro}

\begin{figure*}[t]
  \centering
  \includegraphics[width=\linewidth]{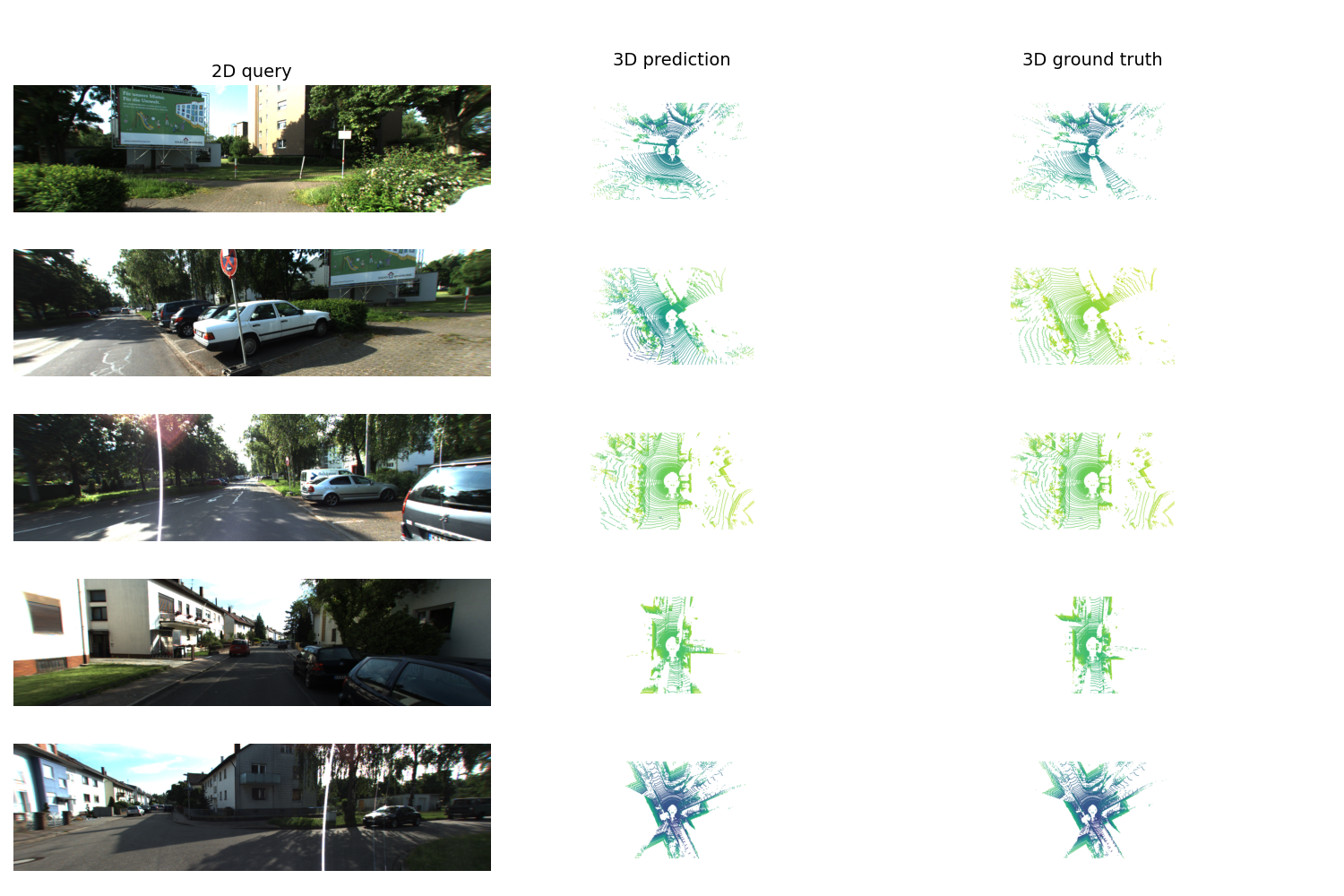}
  \caption{Visualization of 2D to 3D localization}
  \label{fig:2D_3D_visualization}
\end{figure*}

\begin{figure*}[t]
  \centering
  \includegraphics[width=\linewidth]{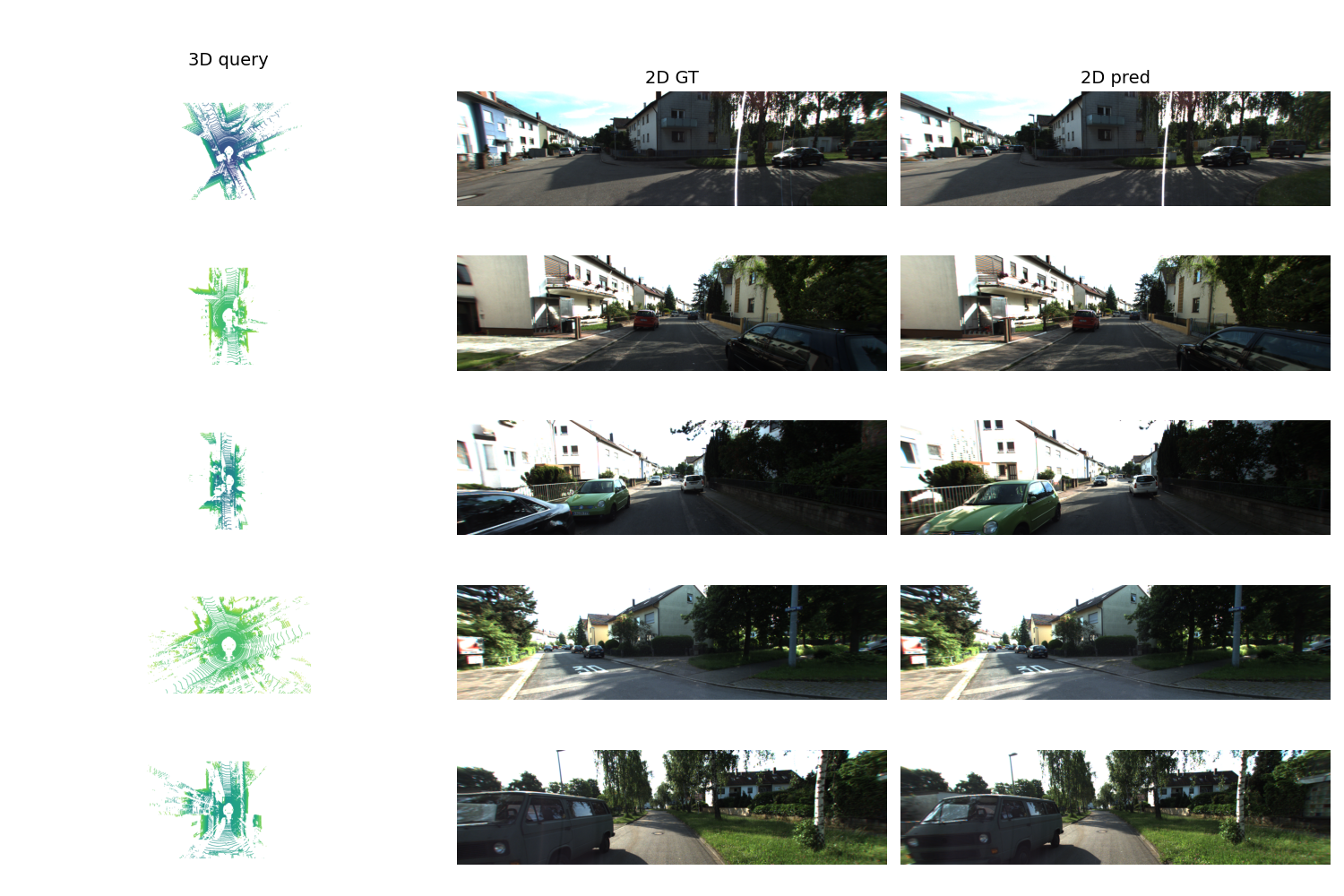}
  \caption{Visualization of 3D to 2D localization}
  \label{fig:3D_2D_visualization}
\end{figure*}

Fig \ref{fig:2D_3D_visualization} and Fig \ref{fig:3D_2D_visualization} demonstrate the qualitative results of 2D to 3D localization and 3D to 2D localization respectively on KITTI-360 dataset using the best model LIP-Loc. The 3D scans are shown in top view. As can be seen in Fig \ref{fig:2D_3D_visualization}, the 3D scans that we are able to predict are very close to ground truth scans, which can help in navigation in an environment where 3D information is not available at test time. Similarly, Fig \ref{fig:3D_2D_visualization} shows that the RGB images retrieved via 3D-2D localization are similar to ground truth, thus this could further result in downstream application like finding a finer pose through perspective-n-point algorithm.

Let us discuss about semantic breakdown of KITTI-360 dataset now. KITTI-360 is a diverse suburban dataset with 37 label classes, including 24 “instance” classes and 13 “stuff” classes. They define a category and within a category come many classes. For example, the category “flat” contains “classes” like road, sidewalk, parking etc; construction contains building, garage, wall, fence etc. They additionally do a statistical analysis over the distribution of the semantic labels, through which they plot 2D semantic labels over frames and 3D semantic labels over points and bounding boxes. When it is done over frames or points, they find that the highest distribution is of classes vegetation, sky, terrain, car and road while when done over bounding boxes reflects that the highest distribution is for classes car, pedestrian, rider, building and bicycle. There are some predominantly “downtown” scenes, i.e. those with buildings/houses, and many objects like trees, bicycles are common, such as sequences 0000, 0002, 0009. There are also some predominantly “highway” scenes, i.e. i.e. those with open areas, continuous vegetation, roads and cars such as 0003, 0004, 0005. Fig-5 of main paper reported recall@1 values on these sequences.  Overall, without any training, our "Zero-shot LIP-Loc" performs well in all sets of diverse conditions having a recall of around 0.5 for most sequences and reaching a maximum of 0.658 for 1 sequence. There is no clear correlation between the accuracy on a sequence and its semantic distribution, i.e. whether it is highway or downtown. For example, if we look at highway scenes such as sequences 0003 and 0005 have recall of 0.658 and 0.594, whereas other highway sequences like 0004 have low recall value like 0.470 whereas downtown scene like 0000 has recall of 0.541. This could mean that our “Zero-shot LIP-Loc” model is not learning spurious correlations, in other words, it is not fitting to certain distribution, rather it is learning in a generalized way. As opposed to our baseline AECMLoc which has tested only on 0000 which has one kind of distribution predominantly, we have tested on 6 sequences each of which differs and we get good recall values for each and do not get abnormally poor values anywhere, which suggests that our approach is robust to distribution shift. With that being said, we have to point out that KITTI-360 does not give clear per-sequence breakdown of semantics, and there is a necessity for a benchmark to do thorough analysis and demonstrate the true zero-shot effectiveness of approaches like ours.

\begin{figure*}[t]
  \centering
  \includegraphics[width=\linewidth]{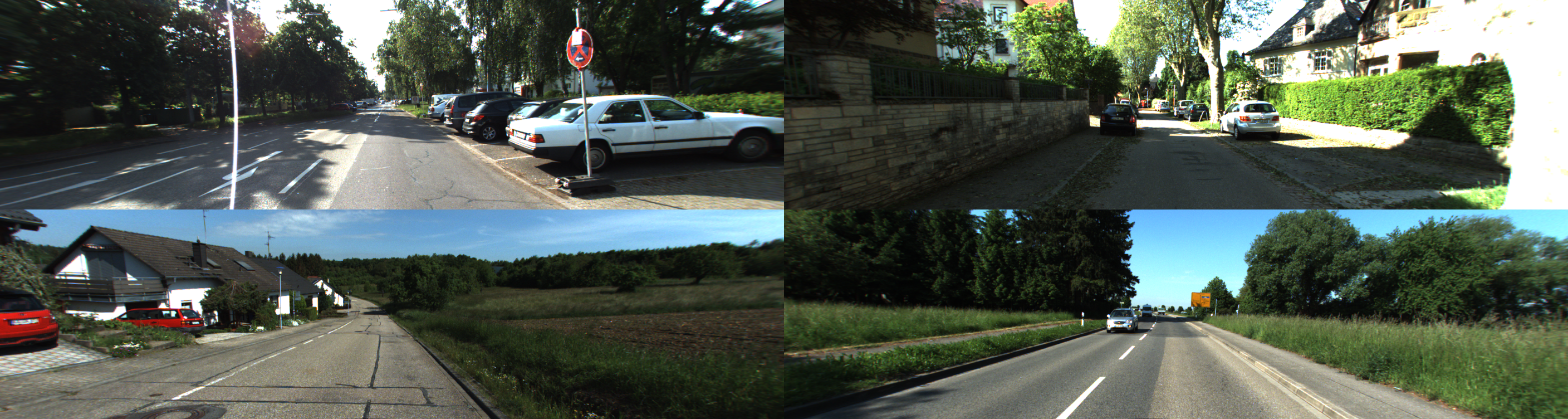}
  \caption{Semantic Breakdown of KITTI-360 evaluation sequences: The top row represents downtown with cars, buildings whereas bottom row represents highway with more greenery, wide roads. Our "Zero-shot LIP-Loc" model performs well in these diverse conditions without even being trained on this data.}
  \label{fig:semantic_breakdown}
\end{figure*}

%-------------------------------------------------------------------------

\begin{figure*}[t]
  \centering
  \includegraphics[width=\linewidth]{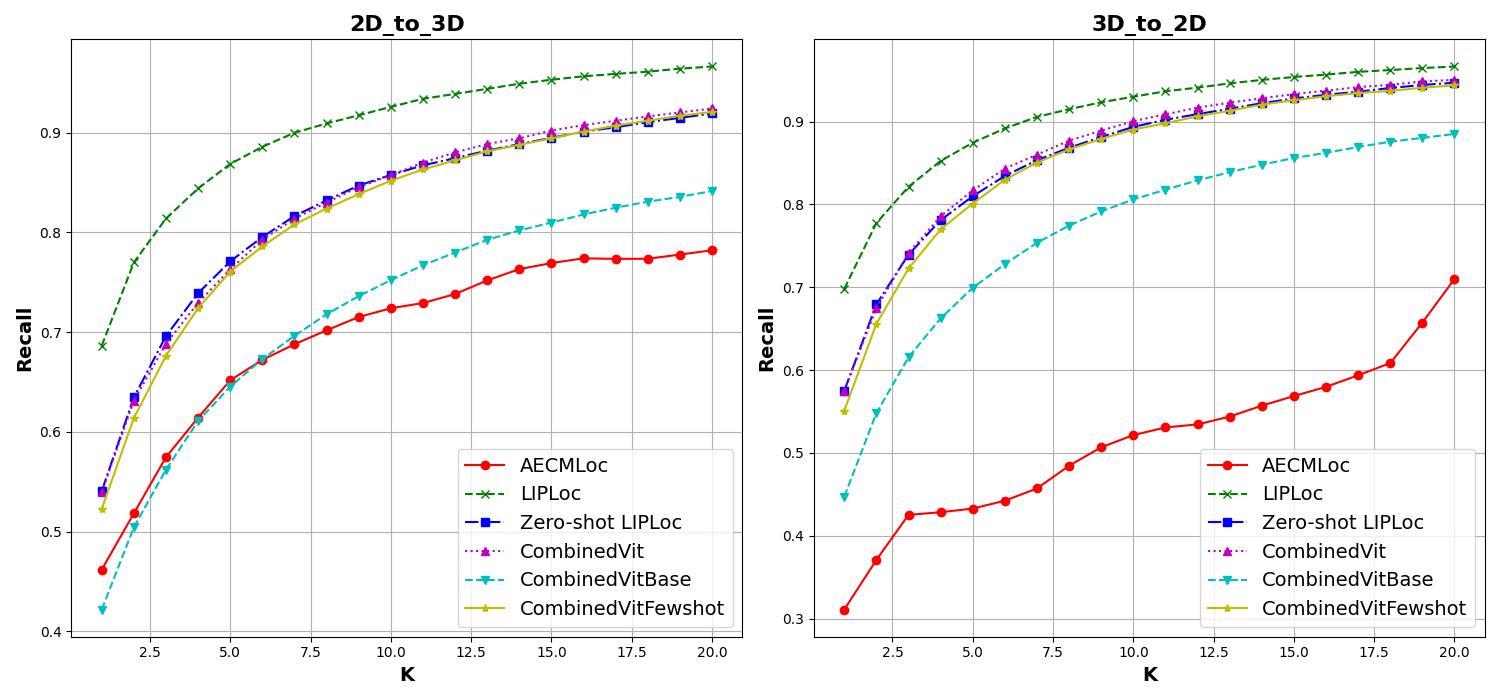}
  \caption{Recall@K curves on KITTI-360 sequence 0000: Combined Models comparison with our best models and baseline AECMLoc}
  \label{fig:aecmloc_comparison_plot}
\end{figure*}

%-------------------------------------------------------------------------
\section{Architecture: Different Encoders \& Bigger Models}
\label{sec:formatting}
\begin{table}[htbp]
\centering
\begin{tabular}{|l|c|c|}
\hline
Different Encoder & Seq 8 & Seq 9 \\
\hline
exp\_large (resnet) & 0.179 & 0.147 \\
\hline
exp\_larger (resnet) & 0.295 & 0.309 \\
\hline
exp\_largest (resnet) & 0.484 & 0.457 \\
\hline
trip\_larger\_vanila (resnet) & 0.215 & 0.232 \\
\hline
exp\_large (ViT) & 0.278 & 0.260 \\
\hline
exp\_larger (ViT) & 0.547 & 0.525 \\
\hline
exp\_largest (ViT) & 0.805 & 0.780 \\
\hline
trip\_vanila\_larger (ViT) & 0.279 & 0.282 \\
\hline
\end{tabular}
\caption{Recall@1 on Different encoders}
\label{tab:diff_encoder}
\end{table}
%-------------------------------------------------------------------------

%-------------------------------------------------------------------------

\begin{table}[htbp]
\centering
\begin{tabular}{|l|c|c|}
\hline
Bigger Model & Seq 8 & Seq 9 \\
\hline
exp\_largest\_resnet101 & 0.477 & 0.462 \\
\hline
exp\_large\_resnet101 & 0.178 & 0.152 \\
\hline
exp\_largest\_vit\_base\_patch16\_224 & 0.777 & 0.720 \\
\hline
exp\_large\_vit\_base\_patch16\_224 & 0.238 & 0.230 \\
\hline
\end{tabular}
\caption{Recall@1 on Bigger Models}
\label{tab:bigger_models}
\end{table}

%-------------------------------------------------------------------------

\begin{table}[htbp]
\centering
\begin{tabular}{|l|c|c|}
\hline
Combined Models & Seq 8 & Seq 9 \\
\hline
exp\_largest (ViT) & 0.805 & 0.780 \\
\hline
exp\_combined\_vit (thresh 50) & 0.817 & 0.741 \\
\hline
exp\_combined\_vit (thresh 100) & 0.785 & 0.758 \\
\hline
exp\_combined\_vit\_fewshot & 0.811 & 0.773 \\
\hline
exp\_combined\_vit\_base\_patch16\_224 & 0.827 & 0.805 \\
\hline
\end{tabular}
\caption{Recall@1 for model trained on both KITTI and KITTI-360 and inference on KITTI-360 (no overlap): test on 0000 of KITTI-360, while train on the rest.}
\label{tab:combined_models}
\end{table}
%-------------------------------------------------------------------------

In this section, we report additional experiments on experimenting with different encoders and bigger models. We report in the main paper that \textit{vit\_small\_patch16\_224} is the final model we have chosen. Here we discuss more on why we have chosen that model and what results we have obtained for other models. Please do note that in this supplementary too, wherever not explicitly mentioned, we are referring to ViT's model of \textit{vit\_small\_patch16\_224} model and ResNet's model of \textit{ResNet50}.

Table \ref{tab:diff_encoder} reports recall values on different encoders. We observed in our experiments that ViT models have a significant accuracy improvement over ResNet. To ensure the comparison is fair, we pick models which have roughly same number of parameters, i.e. \textit{ResNet50} which has 25M parameters (25,557,032) and \textit{vit\_small\_patch16\_224} which has 22M parameters (22,050,664). Even with 3M less parameters, we notice a rise of over 30\% accuracy in \texttt{exp\_largest} case. This pattern can be observed in training over smaller sequences too, such as \texttt{exp\_larger} and \texttt{exp\_large}. Even in the triplet vanilla case, we can see a marginal 5\% improvement, clearly demonstrating that the edge of ViT over the standard ResNet models.

In Table \ref{tab:bigger_models}, we report the recall values of bigger models such as \textit{resnet101} and \textit{vit\_base\_patch16\_224}. Although these models have significantly higher parameters, such as 87M for the latter, we do not observe any much change in accuracy. In fact, it dropped marginally. This could be due to the fact that localization datasets are much smaller compared to internet-scale datasets like CLIP and bigger models result in overparametrization, thus dropping accuracy. 

Does adding more data improve accuracy for these bigger models? In the main paper, we have discussed the standard train-test setting, where we tested on 0000 sequence of KITTI-360 while its training was on rest of sequences of KITTI-360 (0000) or KITTI (Seq 8 and Seq 9), this was "LIP-Loc". Other  setting was when we trained on KITTI data and evaluated on KITTI-360, called as "Zero-shot LIP-Loc". \texttt{exp\_combined} refers to the a third setting, where we train on all sequences of KITTI and KITTI-360 excluding test sequences of KITTI-360 (i.e. 0000) and KITTI (i.e. 8 and 9) on which we test. To get back to our question of whether adding more data will improve accuracy for bigger models, see last row of Table \ref{tab:combined_models} whose accuracy improved over \texttt{exp\_largest\_vit\_base\_patch16\_224} of \ref{tab:bigger_models} by 5\%. This further reaffirms that if we scale the model, we need to scale the data in order to improve the accuracy. Do note that this fact is not as established in visual localization as much as in computer vision or language models, it is still an open question as to how much role big data will play for localization, hence these analyses play crucial role.

We also try a few-shot experiment here wherein we give just 1\% of data of KITTI-360 when compared to the \texttt{exp\_combined\_vit} experiment. To be clear, \texttt{exp\_combined\_vit} uses all sequences of KITTI and KITTI-360 (excluding test sequences), whereas \texttt{exp\_combined\_vit\_fewshot} uses all sequences of KITTI but just 1\% of KITTI-360. This is a very captivating result: We receive almost same or marginally improve upon the accuracy as the other experiment despite using significantly very less dataset.

It is also worth noting that the combined experiments don't improve significantly from \texttt{exp\_largest}, unless we use a bigger model like \texttt{exp\_largest\_vit\_base\_patch16\_224}, which is also 2\% improvement. Future experiments have to be done to establish even clearer understanding.

So far, we have discussed evaluation on KITTI dataset. Now let us discuss about evaluation on KITTI-360 dataset by looking at Fig \ref{fig:aecmloc_comparison_plot}. Previously in the main paper, we reported LIPLoc, Zero-shot LIPLoc and AECMLoc (baseline). Here we additionally add the plots of combined experiments, which as described above, merges the training sequences of KITTI and KITTI-360 and trains a single model using full data. Do note that all of our models beat the SOTA AECMLoc. But amongst our models themselves, we rather see ambigious or counterintuitive results. 

Firstly to clarify, when we use the term "LIP-Loc", we are referring to standard train/test paradigm, for example when reporting LIP-Loc on KITTI-360, we mean we trained on certain train split of KITTI-360 and evaluating on its test splits; similarly when reporting on KITTI, we mean we trained on train split of KITTI and evaluating on its corresponding test splits. "Zero-shot LIP-Loc" on other hand are trained on full KITTI data but has not seen any KITTI-360 data on which we evaluate. Whereas combined models are trained on train split of KITTI and train split of KITTI-360. Therefore, please keep these nuances in mind when interpreting the result. With that being said, since we are evaluating on test split of KITTI-360, we would expect combined models to significantly outperform Zero-shot LIP-Loc. However, that's not the case here: Amongst the combined models, all the standard ViT model \texttt{CombinedVit} and bigger model \texttt{CombinedVitBase} and the few shot model \texttt{CombinedVitFewshot} give similar recall compared to Zero-shot LIP-Loc and subpar performance compared to LIP-Loc. This further proves that Zero-shot LIP-Loc has generalized very well. 

As future work, it will be interesting to see an analysis between zero-shot and few-shot LIP-Loc. This raises many open questions: In computer vision problems which CLIP deals with, few-shot is clearly defined because it is talking about classification categories. However it is not well defined in visual localization context, which further asserts the necessity of establishment of a well thought benchmark. We encourage the reader to address these open questions and ask the question, "Can big data solve the localization problem?"

CLIP admits that it is not good at task generalization for tasks such as finding close objects in an image or counting the number of objects in an image. Extending our work along the lines of the recent work LiDARCLIP~\cite{hess2023lidarclip} which connects CLIP's embedding space to LiDAR point cloud domain could result in an approach which uses text features to query the right set of points in the LiDAR scan, explicitly identify distance and location of the objects and applying clustering in 3D space to count number of objects (for example) and correlate them with image features to identify the class and appearance of an object. This is especially helpful in extreme low visibility conditions where RGB camera will not work well and LiDAR can help identify objects close to the ego vehicle.

\section{Architecture: Hierarchical Design}
\label{sec:formatting}

In models as a follow up to CLIP, many models such as ViCHA~\cite{vicha_shukor2022efficient} propose architectural improvement such as hierarchical alignment. What this essentially proposes is that aligning the two encoders at various levels by adding multiple losses at various layers of text and image encoder. They claim that this helps in convergence faster and results in superior performance. In our experiments, we have hierarchically aligned image and lidar encoders at various layers and report it in first half of the table \ref{tab:architecture_hier_design}. We have tried two experiments: One that aligns only at final layers, the other that aligns throughout the encoder, as ViCHA argues that aligning at the beginning could result in confusing the model. However, in our experiments we did not observe any noticeable improvement, although ViCHA's observation of alignment at final layers could be verified in the case of visual localization as well.

The second half of the table pertains to the following. In standard CLIP setting, there is no relation between any consecutive images in a batch, as they are just (image, text) pairs. However, in our localization setting, the images are sequential. Therefore, we attempted the question: Can we achieve higher accuracy by grouping together adjacent images and having additional encoder for groups of images which results in secondary loss? The last 3 rows of Table \ref{tab:architecture_hier_design} correspond to these experiments. Do note that these experiments are with \textit{ResNet} architecture. Our results actually deteriorated during our experiments. There is a simple rationale for this: The training of deep models works so well because of randomization of samples in a batch, especially in the case of our batch construction technique. When we contruct groups within the batch and ensure the images within the group are consecutive but groups themselves are random, we are asking for a tradeoff: will the additional hierarchical loss improve accuracy more than reducing randomization will decrease it? We have found in our experiments that the answer is no, even for smalller group sizes such as 4. 

This section concludes that sticking to non-complicated architectures works the best since the power of CLIP model subsumes any minor architectural improvement.

%-------------------------------------------------------------------------
\begin{table}[htbp]
\centering
\begin{tabular}{|l|c|c|}
\hline
Advanced Architectures & Seq 8 & Seq 9 \\
\hline
exp\_large (ViT) & 0.278 & 0.260 \\
\hline
hier\_align\_large\_vit (final layers) & 0.275 & 0.258 \\
\hline
hier\_align\_large\_vit (all layers) & 0.218 & 0.197 \\
\hline
exp\_large (resnet) & 0.179 & 0.147 \\
\hline
exp\_larger (resnet) & 0.295 & 0.309 \\
\hline
exp\_largest (resnet) & 0.484 & 0.457 \\
\hline
hier\_group\_shuffle\_large\_resnet & 0.170 & 0.1495 \\
\hline
hier\_group\_shuffle\_larger\_resnet & 0.239 & 0.212 \\
\hline
hier\_group\_shuffle\_largest\_resnet & 0.378 & 0.346 \\
\hline
\end{tabular}
\caption{Architecture: Hierarchical Design}
\label{tab:architecture_hier_design}
\end{table}

%-------------------------------------------------------------------------

%-------------------------------------------------------------------------

%%%%%%%%% REFERENCES
% {\small
% \bibliographystyle{ieee_fullname}
% \bibliography{egbib}
% }

% \end{document}

\end{document}